\documentclass[conference]{IEEEtran}
\IEEEoverridecommandlockouts

\usepackage{cite}
\usepackage{amsmath,amssymb,amsfonts}
\usepackage{algorithmic}
\usepackage{graphicx}
\usepackage{textcomp}
\usepackage{xcolor}
\usepackage{hyperref}
\usepackage{makecell}
\usepackage{booktabs}
\usepackage{multirow}
\usepackage{colortbl}

\def\BibTeX{{\rm B\kern-.05em{\sc i\kern-.025em b}\kern-.08em
    T\kern-.1667em\lower.7ex\hbox{E}\kern-.125emX}}

\definecolor{R1}{rgb}{0.720, 0.850, 0.600} 
\definecolor{R2}{rgb}{0.800, 0.750, 0.600} 
\definecolor{R3}{rgb}{0.870, 0.670, 0.500} 
\definecolor{R4}{rgb}{0.850, 0.570, 0.450} 
\definecolor{R5}{rgb}{0.820, 0.450, 0.380} 

\begin{document}

\title{CSS: Overcoming Pose and Scene Challenges in Crowd-Sourced 3D Gaussian Splatting\\
}

\author{\IEEEauthorblockN{Runze Chen\textsuperscript{1,2*}, Mingyu Xiao\textsuperscript{1,2*}, Haiyong Luo\textsuperscript{2$\dagger$}, Fang Zhao\textsuperscript{1$\dagger$}, Fan Wu\textsuperscript{1,2}, Hao Xiong\textsuperscript{1,2}, Qi Liu\textsuperscript{3}, Meng Song\textsuperscript{3}}
\IEEEauthorblockA{
\textsuperscript{1}\textit{Beijing University of Posts and Telecommunications, Beijing, China}\\
\textsuperscript{2}\textit{Institute of Computing Technology, Chinese Academy of Sciences, Beijing, China}\\
\textsuperscript{3}\textit{China Unicom Smart City Research Institute, Beijing, China}\\
\{chenrz925,shawnmy,zfsse,wufan98326,xmr2015211989\}@bupt.edu.cn, yhluo@ict.ac.cn, \{liuqi49,songmeng\}@chinaunicom.cn}
\thanks{* These authors contributed equally to this work.}
\thanks{$\dagger$ Corresponding authors: Haiyong Luo and Fang Zhao.}
}

\maketitle

\begin{abstract}
We introduce Crowd-Sourced Splatting (CSS), a novel 3D Gaussian Splatting (3DGS) pipeline designed to overcome the challenges of pose-free scene reconstruction using crowd-sourced imagery. The dream of reconstructing historically significant but inaccessible scenes from collections of photographs has long captivated researchers. However, traditional 3D techniques struggle with missing camera poses, limited viewpoints, and inconsistent lighting. CSS addresses these challenges through robust geometric priors and advanced illumination modeling, enabling high-quality novel view synthesis under complex, real-world conditions. Our method demonstrates clear improvements over existing approaches, paving the way for more accurate and flexible applications in AR, VR, and large-scale 3D reconstruction.
\end{abstract}


\section{Introduction}
\label{sec:intro}

Reconstructing historically significant or inaccessible scenes from existing crowdsourced or archival photographs \cite{doi:10.1177/14780771231168224,22d9af9531054c88a8a46c194f6b3712,Magnani_Douglass_Schroder_Reeves_Braun_2020} has long been a key objective in fields such as virtual reality (VR), augmented reality (AR), and autonomous driving \cite{survey}. While 3D Gaussian Splatting (3DGS) \cite{3dgs} has made significant strides in achieving high-fidelity, real-time rendering through its differentiability, its primary focus is on optimized visual representation rather than the reconstruction of scenes from diverse, unstructured image sources \cite{10630672}. In contrast, crowdsourcing has emerged as a transformative approach for data aggregation in 3D visual computation, significantly lowering the cost and time required for data collection compared to traditional methods \cite{10.1145/3569090}. By harnessing the diversity and widespread availability of crowdsourced imagery, researchers can achieve more comprehensive and varied datasets, laying the groundwork for more detailed and nuanced 3D reconstructions \cite{csnerf}.

Crowdsourced imagery presents unique challenges that complicate the construction of 3DGS models. The primary issues include the lack of precise camera poses \cite{colmapfree}, sparse and limited viewpoints \cite{10446844}, and inconsistent lighting conditions across images \cite{wildinnerf}. These challenges are further compounded by the absence of positional priors, and the temporal and spatial variations in lighting \cite{9413356} caused by the asynchronous nature of crowdsourced data. Such inconsistencies disrupt traditional methods like COLMAP \cite{colmap1,colmap2}, which rely on accurate Structure from Motion (SfM) \cite{DBLP:journals/corr/abs-2312-07504,fan2024instantsplat,10446550}, and they particularly affect the ability of novel view synthesis methods to maintain consistent color and texture across perspectives \cite{Gao_2024_CVPR}. As a result, synthesizing accurate and visually coherent new viewpoints from crowdsourced data remains a significant challenge \cite{DBLP:journals/jair/MogadalaKK21}. Addressing these issues requires more robust approaches, such as leveraging large visual models \cite{mast3r}, which can better generalize across diverse and noisy inputs to enhance pose estimation \cite{Wang_2024_CVPR} and illumination consistency \cite{10480646}.

To address these challenges, we introduce Crowd-Sourced Splatting (CSS), a novel pose-free 3DGS generation pipeline designed for crowd-sourced imagery. The key innovations of CSS include: (1) \textbf{A robust initialization mechanism} utilizing expert models and extensive 2D geometric priors to overcome the lack of precise camera poses and inconsistent imaging conditions in crowdsourced data. (2) A\textbf{n advanced illumination model} employing high-order spherical harmonics to harmonize varying lighting conditions and perspectives, ensuring consistent and high-quality 3DGS under complex crowdsourced scenarios. (3) \textbf{The development of CSScenes}, a comprehensive dataset sourced from internet-based crowd imagery, providing benchmarks across diverse indoor and outdoor environments.

These innovations mark a major advancement in 3D visual computation, directly addressing the key challenges of crowdsourced imagery. By eliminating the need for precise camera poses and incorporating advanced illumination modeling, CSS generates high-quality 3DGS models even in complex and varied conditions. The introduction of CSScenes strengthens the practical impact of our approach, providing a valuable benchmark for future research and development in this field.

\section{Methodology}
\label{sec:method}

\begin{figure*}
    \centering
    \includegraphics[width=0.8\linewidth]{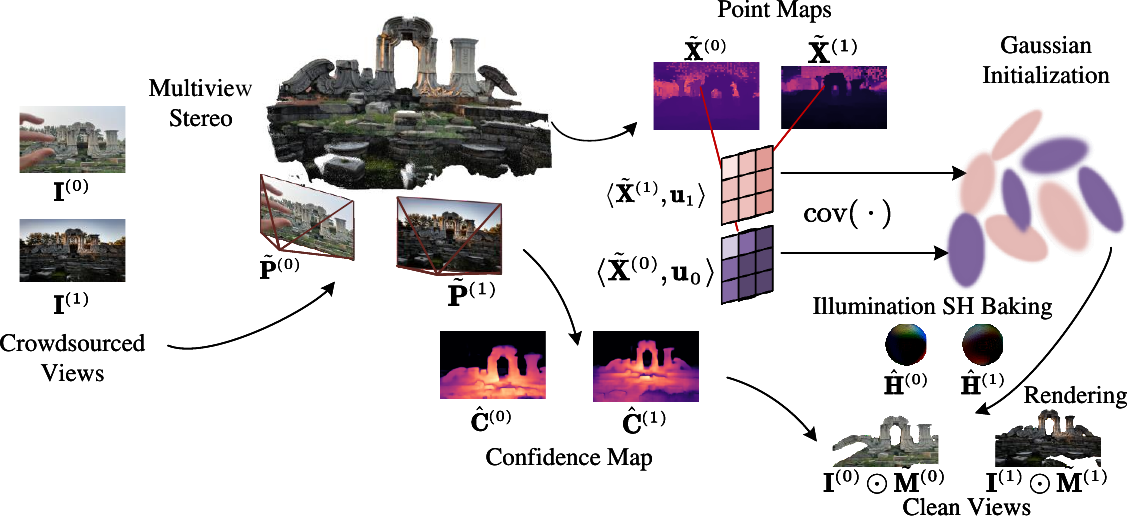}
    \caption{\textbf{CSS computational pipeline.} We employ multiview stereo estimation to determine the orientation of each crowdsourced viewpoint \( \tilde{\mathbf{P}}^{(i)} \), alongside a confidence map \( \hat{\mathbf{C}}^{(i)} \) and a corresponding point cloud \( \tilde{\mathbf{X}}^{(i)} \). The covariance \( \mathrm{cov}(\langle \tilde{\mathbf{X}}^{(1)},\mathbf{u}_1\rangle ) \) is calculated using the adjacent points within the point cloud to initialize the Gaussian distribution. Throughout the 3D Gaussian refinement process, we model the illumination from each crowdsourced viewpoint \( i \) using higher-order spherical harmonics, which allows us to render the scene effectively and construct a stable and coherent novel viewpoint synthesis.
}
    \label{fig:overview}
\end{figure*}

To tackle the challenges of missing pose information and significant lighting variations in the complex task of crowdsourced 3DGS reconstruction, we propose an innovative pipeline within the CSS framework. This pipeline allows for the synthesis of novel viewpoints from diverse and challenging crowdsourced data. Figure \ref{fig:overview} illustrates the overall structure of the proposed CSS pipeline.

\subsection{Robust Initialization for Crowdsourced 3DGS Reconstruction}
\label{sec:rob_init}

The sparse viewpoints in crowdsourced imagery and the diverse configurations of capture devices present significant challenges to recovering accurate camera poses. To address these challenges, we leverage the geometric and structural priors provided by MASt3R \cite{mast3r} to initialize the 3DGS reconstruction.

Given an image pair $(\textbf{I}^{(i)}, \mathbf{I}^{(j)})$ within the crowdsourced image set $\mathcal{I}$, we derive the corresponding 3D point maps $(\hat{\mathbf{X}}^{(i)}, \hat{\mathbf{X}}^{(j)})$ and dense feature maps $(\mathbf{F}^{(i)}, \mathbf{F}^{(j)})$. These feature maps, denoted as $\mathbf{D}^{(i)}$ and $\mathbf{D}^{(j)}$, capture the geometric and texture characteristics of each pixel robustly. We then perform fast reciprocal matching to identify stable pairs of corresponding pixels, denoted as \(\mathcal{R}^{(i,j)} = \{(\mathbf{u}^{(i)}, \mathbf{u}^{(j)})\}\). Here, \(\mathbf{u}^{(i)}\) and \(\mathbf{u}^{(j)}\) are pixel coordinates that minimize the distance between their respective feature descriptors \(\mathbf{D}^{(i)}_{\mathbf{u}^{(i)}}\) and \(\mathbf{D}^{(j)}_{\mathbf{u}^{(j)}}\). These matches are critical for robust camera pose estimation.

Using MASt3R, we estimate initial 3D point maps \(\hat{\mathbf{X}}^{(i)}\), which are weighted by confidence maps \(\hat{\mathbf{C}}^{(i)}\) to estimate initial camera intrinsics \(\hat{\mathbf{K}}^{(i)}\) for each view. Subsequently, we refine these estimates, resulting in optimized camera intrinsics \(\tilde{\mathbf{K}}^{(i)}\), 3D point coordinates \(\tilde{\mathbf{X}}^{(i)}\), and camera extrinsics \(\tilde{\mathbf{P}}^{(i)}\), using a coarse-to-fine joint optimization process. During optimization, the predictions \(\hat{\mathbf{X}}^{(i)}\) derived from all image pairs \((\mathbf{I}^{(i)}, \mathbf{I}^{(j)})\) are used to iteratively refine \(\tilde{\mathbf{P}}^{(i)}\) and \(\tilde{\mathbf{X}}^{(i)}\) for each view \(i\).

The optimization process employs a distance-based loss function \(\mathcal{L}_\mathrm{D}\) to improve the estimates for each perspective:
\begin{equation}
    \mathcal{L}_\mathrm{D}=\frac{1}{\sum\limits_{\mathcal{R}^{(i,j)}}\hat{\mathbf{C}}^{(i)}}\sum_{\mathcal{R}^{(i,j)}}{\hat{\mathbf{C}}^{(i)}\|\tilde{\mathbf{X}}^{(i)}-\tilde{\mathbf{P}}^{(j)} \tilde{\mathbf{P}}^{(i)^{-1}} \hat{\mathbf{X}}^{(i)}\|^2},
\end{equation}
where \(\tilde{\mathbf{X}}^{(i)}\) and \(\tilde{\mathbf{P}}^{(i)}\) are the optimized estimates for the 3D points and camera extrinsics, while \(\hat{\mathbf{X}}^{(i)}\) and \(\hat{\mathbf{P}}^{(i)}\) are the initial estimates from MASt3R.

During the coarse optimization phase, the camera extrinsics are optimized by minimizing the 3D distances between all matched point pairs \(\mathcal{R}^{(i,j)}\):
\begin{equation}
    \mathcal{L}_\mathrm{C}=\frac{1}{\sum\limits_{\mathcal{R}^{(i,j)}}\hat{\mathbf{C}}^{(i)}}\sum_{\mathcal{R}^{(i,j)}}{\hat{\mathbf{C}}^{(i)}\|\hat{\mathbf{X}}^{(i)}-\tilde{\mathbf{P}}^{(j)} \tilde{\mathbf{P}}^{(i)^{-1}} \hat{\mathbf{X}}^{(i)}\|^2},
\end{equation}
where \(\hat{\mathbf{X}}^{(i)}\) and \(\hat{\mathbf{P}}^{(i)}\) are the initial estimates from MASt3R. The final loss function in the coarse optimization phase is defined as \(\mathcal{L}_\mathrm{S1} = \mathcal{L}_\mathrm{D} + \lambda\mathcal{L}_\mathrm{C}\), where \(\lambda\) is a weighting factor.

To ensure accurate reconstruction for novel view synthesis, we enhance the reprojection accuracy for each view \(i\) through fine-grained optimization. This phase uses the following loss function:
\begin{equation}
    \mathcal{L}_\mathrm{F}=\frac{1}{\sum\limits_{\mathcal{R}^{(i,j)}}\hat{\mathbf{C}}^{(i)}}\sum_{\mathcal{R}^{(i,j)}}{\hat{\mathbf{C}}^{(i)}\|\mathbf{u}^{(i)}-\pi(\tilde{\mathbf{K}}^{(i)},\tilde{\mathbf{P}}^{(i)},\tilde{\mathbf{X}})\|^2},
\end{equation}
where \(\mathbf{u}^{(i)}\) denotes the 2D pixel coordinates in view \(i\), and \(\pi(\cdot)\) is the projection function utilizing the optimized camera intrinsics \(\tilde{\mathbf{K}}^{(i)}\), extrinsics \(\tilde{\mathbf{P}}^{(i)}\), and 3D point coordinates \(\tilde{\mathbf{X}}\). The fine-grained loss in this phase is given by \(\mathcal{L}_\mathrm{S2} = \mathcal{L}_\mathrm{F} + \lambda\mathcal{L}_\mathrm{C}\), where \(\lambda\) is the weighting factor.

As illustrated in Figure \ref{fig:overview}, we initialize the covariance of the 3DGS by leveraging the inherent 3D geometric relationships present in the point map. For each point \(u\) in the point map \(\tilde{\mathbf{P}}^{(i)}\), denoted as \(\tilde{\mathbf{P}}^{(i)}_u\), we define a local \(3 \times 3\) neighborhood \(\langle \tilde{\mathbf{X}}^{(i)},\mathbf{u}\rangle \in \mathbb{R}^{(3\times 3)\times3}\). The local neighborhood point set \(\langle \tilde{\mathbf{X}}^{(i)},\mathbf{u}\rangle\) undergoes singular value decomposition (SVD):
\begin{equation}
    \mathrm{cov}(\langle \tilde{\mathbf{X}}^{(i)},\mathbf{u}\rangle) = \mathbf{U} \mathbf{S}^2 \mathbf{V}^T,
\end{equation}
where \(\mathbf{U}\) and \(\mathbf{V}\) are orthogonal matrices representing the left and right singular vectors of the covariance matrix $\mathrm{cov}(\langle \tilde{\mathbf{X}}^{(i)},\mathbf{u}\rangle)$, and \(\mathbf{S}^2\) is a diagonal matrix containing the singular values. The diagonal elements of \(\mathbf{S}\) define the scale transformation of the 3DGS, while \(\mathbf{V}\) specifies the rotational transformation.

Accurately predicting depth is one of the most challenging aspects of estimating 3D coordinates from images. Errors in depth estimation can lead to overestimation of Gaussian scales, causing rendering failures. To address this, we regularize the largest component of \(\mathbf{S}\) with a clipping function, yielding a new scale transformation:
\begin{equation}
    \mathbf{S}^\prime = \mathrm{clip}(\mathbf{S}, \mathrm{median}(\mathbf{S}), \min(\mathbf{S})),
\end{equation}
where \(\mathrm{clip}\) constrains the values of \(\mathbf{S}\) within the range set by its median and minimum values, thus preventing excessive scaling that could impair the rendering process. The covariance matrix of the initialized 3DGS is then given by $\mathbf{\Sigma} = \mathbf{U}\mathbf{S^\prime}^2 \mathbf{V}^T$. This normalized initialization offers a robust foundation for further refinement of the 3DGS.

\subsection{Refinement of 3DGS to Mitigate Illumination and Dynamic Biases}

Crowdsourced imagery, affected by factors like time, weather, and dynamic objects, often causes occlusions and lighting variations in the target scene, creating challenges for the 3DGS training pipeline. To address these issues, we designed the CSS pipeline to account for the distribution of occlusions and lighting variations during training.

\textbf{Occlusions.} Occlusions occur when dynamic objects, such as people or vehicles, obstruct the camera’s view, making parts of the target scene hidden or partially visible. These occlusions introduce significant challenges in 3DGS training, resulting in inconsistencies across images and a loss of crucial information, ultimately affecting the accuracy and completeness of the reconstructed scenes. For each crowdsourced view \(i\), we estimate the confidence map \(\hat{\mathbf{C}}^{(i)}\) using a multiview approach and apply a threshold to distinguish occluded regions with mask $\tilde{\textbf{M}}^{(i)}$. Alternatively, for more focused, compact scenes such as statues or artifacts, the Otsu method \cite{otsu} can be employed to determine the scene's regions automatically.

\textbf{Illumination variations.} The biggest challenge with illumination variation in crowdsourced imagery is the inconsistency in lighting conditions across different images. Since the images are taken at different times and under various lighting environments, like sunlight or artificial light, it causes changes in shadows, brightness, and colors. This makes it harder to match features, estimate depth, and build accurate 3D models, as objects can look very different in each image. According to Retinex theory \cite{retinex}, we can decompose the illumination component \(\mathbf{L}^{(i)}\), influenced by varying lighting environments (e.g., day and night conditions, natural versus artificial light, or different weather scenarios), from the invariant color and texture representation, the reflectance component \(\mathbf{R}^{(i)}\). One of the greatest challenges in applying 3DGS to crowdsourced imagery is the variation in the distribution of illumination components across different views. This insight leads us to a strategy: by expressing the image as \(\mathbf{I}^{(i)} = \mathbf{L}^{(i)} \odot \mathbf{R}^{(i)}\), where \(\odot\) denotes element-wise multiplication, we can isolate the constant reflectance component across views, thereby enhancing the robustness of 3DGS under diverse lighting conditions in crowdsourced imagery.

For each 3D point \(\tilde{\mathbf{X}}^{(i)}\), we compute its direction vector as the unit vector \( \mathbf{d} = \tilde{\mathbf{X}}^{(i)} / \|\tilde{\mathbf{X}}^{(i)}\| \), which is then converted to spherical coordinates \( \theta^{(i)} \) and \( \phi^{(i)} \). To model the environmental illumination \(\mathbf{L}^{(i)}\) using spherical harmonics (SH), we define the illumination function as
\begin{equation}
    L(\theta, \phi) = \mathrm{softplus}\left(\sum_{\ell=0}^{L} \sum_{m=-\ell}^{\ell} c_{\ell m} Y_\ell^m(\theta, \phi)\right),
\end{equation}
where \( c_{\ell m} \) are the SH coefficients, and \( Y_\ell^m(\theta, \phi) \) are the SH basis functions, capturing the illumination contribution from different directions. During training, we optimize these SH coefficients via gradient descent, ensuring they accurately represent the illumination distribution for each view. We further employ SH illumination baking to precompute lighting information, allowing us to efficiently capture dynamic lighting variations. Higher-order harmonics, up to order \(10\), are used to capture fine details of the lighting environment and improve the accuracy of the baked illumination.

We utilize 3D Gaussian splatting to render the invariant reflectance component \(\mathbf{R}^{(i)}\), expressed as
\begin{equation}
    \mathbf{R}^{(i)} = \sum_{j=1}^{N} w_j G(\mathbf{u}; \mu_j, \Sigma_j),
\end{equation}
where \( w_j \) represents the reflectance weight of the \( j \)-th Gaussian splat, and \( G(\mathbf{u}; \mu_j, \Sigma_j) \) is the Gaussian function describing the 2D projection of the splat with mean \(\mu_j\) and covariance \(\Sigma_j\) in the image space. During the training phase, we fit the rendered result \(\mathbf{L}^{(i)} \odot \mathbf{R}^{(i)}\) to \(\mathbf{I}^{(i)} \odot \tilde{\mathbf{M}}^{(i)}\), where \(\tilde{\mathbf{M}}^{(i)}\) is derived from the confidence map \(\tilde{\mathbf{C}}^{(i)}\). Here, \(\mathbf{L}^{(i)}\) represents the illumination component, \(\mathbf{I}^{(i)}\) is the original image, and \(\odot\) denotes element-wise multiplication.

\subsection{Crowdsourced 3DGS Data Collection}

Social media platforms and travel websites provide abundant open-licensed, user-contributed multi-perspective imagery. We used destination-specific keywords to extract relevant images, then screened for visual overlap to create a diverse but sparse set for each scene. For other datasets, COLMAP primarily provided reference poses, ensuring consistent pose estimation and reliable evaluation for novel viewpoint synthesis.

\section{Experiments}

\begin{figure*}
    \centering
    \includegraphics[width=0.9\linewidth]{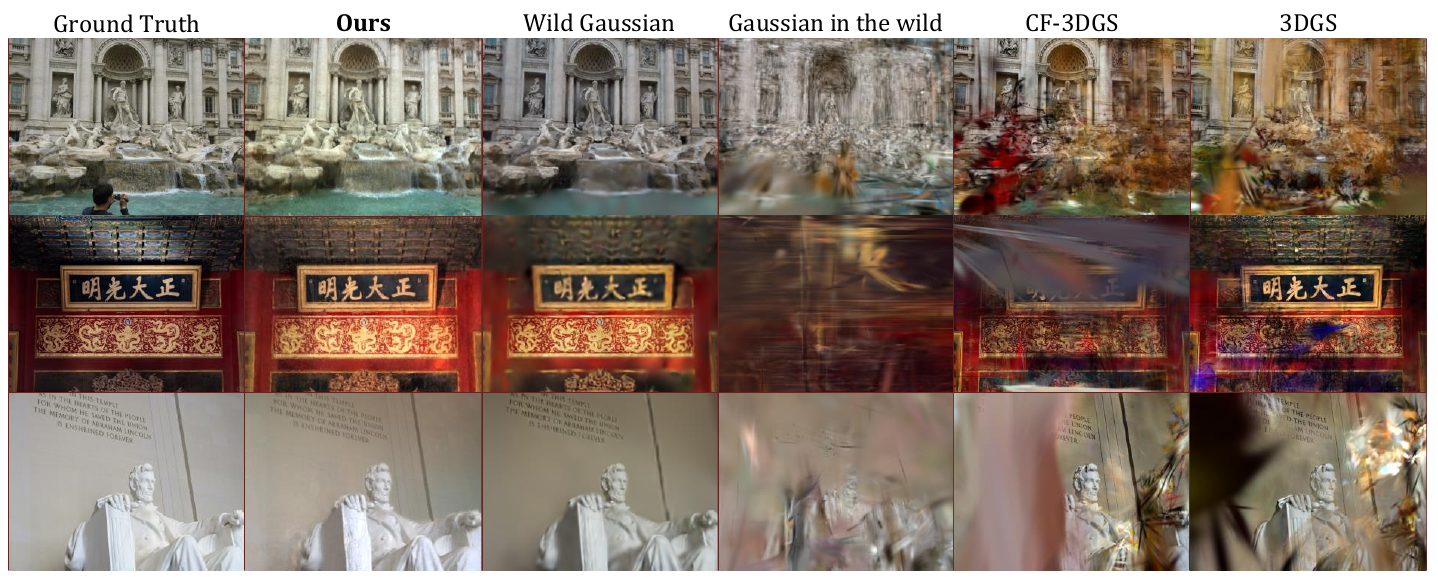}
    \caption{\textbf{Comparison of object appearance rendering across different methods.} Despite varying lighting conditions due to crowd-sourced views, our method achieves more accurate structural and texture preservation than others.}
    \label{fig:render-vis}
\end{figure*}

\begin{table*}[thbp]
    \renewcommand\arraystretch{0.9}
    \tabcolsep=2pt
    \centering
    \caption{\textbf{Quantitative comparison of rendering quality with baselines and ablation study.} Metrics marked with $\uparrow$ favor higher values, while $\downarrow$ prefer lower values. We report the deviation from the full CSS performance for the ablation section. (CM: Confidence Mask, IB: Illumination SH Baking. Ranking from highest to lowest: \colorbox{R1}{\ }\colorbox{R2}{\ }\colorbox{R3}{\ }\colorbox{R4}{\ }\colorbox{R5}{\ })} 
    \label{tab:quant}
    \begin{tabular}{c|ccc|ccc|ccc|ccc|ccc}
        \toprule
        \multirow{4}{*}{Methods} & \multicolumn{9}{c|}{\makecell{CSScenes}} & \multicolumn{6}{c}{\makecell{Photo Tourism}} \\ \cline{2-16}
        & \multicolumn{3}{c|}{\makecell{Bingling Temple\\Sculpture}} & \multicolumn{3}{c|}{\makecell{Qianqing Palace}} & \multicolumn{3}{c|}{\makecell{Yuanmingyuan\\Fountain}} & \multicolumn{3}{c|}
        {\makecell{Lincoln Memorial}} & \multicolumn{3}{c}
        {\makecell{Trevi Fountain}}   \\
        & SSIM$\uparrow$ & PSNR$\uparrow$ & LPIPS$\downarrow$ & SSIM$\uparrow$ & PSNR$\uparrow$ & LPIPS$\downarrow$ & SSIM$\uparrow$ & PSNR$\uparrow$ & LPIPS$\downarrow$ & SSIM$\uparrow$ & PSNR$\uparrow$ & LPIPS$\downarrow$ & SSIM$\uparrow$ & PSNR$\uparrow$ & LPIPS$\downarrow$\\
        \midrule
        3DGS  &\cellcolor{R2}0.5683 &\cellcolor{R2}23.42 &\cellcolor{R3}0.3584 &\cellcolor{R3}0.4215 &\cellcolor{R4}20.91 &\cellcolor{R2}0.3597  &\cellcolor{R1}0.5155 &\cellcolor{R3}21.16 &\cellcolor{R3}0.4755 &\cellcolor{R4}0.8309 &\cellcolor{R4}23.53 &\cellcolor{R4}0.3768 &\cellcolor{R3}0.4715 &\cellcolor{R4}19.20 &\cellcolor{R3}0.5363\\
        Gaussian in the wild &\cellcolor{R5}0.2653 &\cellcolor{R5}19.26 &\cellcolor{R4}0.6679 &\cellcolor{R5}0.2052 &\cellcolor{R5}19.56 &\cellcolor{R4}0.7356 &\cellcolor{R5}0.3583 &\cellcolor{R5}18.82 &\cellcolor{R5}0.7368&\cellcolor{R5}0.6416 &\cellcolor{R5}23.45 &\cellcolor{R5}0.6397 &\cellcolor{R5}0.2398 &\cellcolor{R3}19.99 &\cellcolor{R5}0.6009 \\
        Wild Gaussian &\cellcolor{R4}0.5017 &\cellcolor{R4}22.09 &\cellcolor{R5}0.6915 &\cellcolor{R2}0.4299 &\cellcolor{R2}22.53 &\cellcolor{R5}0.7451 &\cellcolor{R3}0.5002 &\cellcolor{R2}21.84 &\cellcolor{R4}0.6787&\cellcolor{R3}0.8628 &\cellcolor{R2}28.43 &\cellcolor{R3}0.3676   &\cellcolor{R1}0.5171 &\cellcolor{R2}21.62 &\cellcolor{R2}0.5291 \\
        CF-3DGS &\cellcolor{R3}0.5675 &\cellcolor{R3}23.32 &\cellcolor{R2}0.3509 &\cellcolor{R4}0.4168 &\cellcolor{R3}20.96 &\cellcolor{R3}0.4138 &\cellcolor{R4}0.4887 &\cellcolor{R4}20.81 &\cellcolor{R2}0.4745&\cellcolor{R1}0.8677 &\cellcolor{R3}25.59 &\cellcolor{R2}0.3195 &\cellcolor{R4}0.4627 &\cellcolor{R5}18.04 &\cellcolor{R4}0.5434 \\
        Ours&\cellcolor{R1}0.5735&\cellcolor{R1}23.46&\cellcolor{R1}0.3498 &\cellcolor{R1}0.4925&\cellcolor{R1}24.12&\cellcolor{R1}0.2939 &\cellcolor{R2}0.5036 &\cellcolor{R1}22.35 &\cellcolor{R1}0.4511&\cellcolor{R2}0.8660&\cellcolor{R1}29.90&\cellcolor{R1}0.2265  &\cellcolor{R2}0.5167&\cellcolor{R1}23.02&\cellcolor{R1}0.2910\\
        \midrule
        Ours w/o CM&-0.1415&-0.15&+0.0264&-0.1032&-0.49&+0.1601&-0.0986&-0.79&+0.0316&-0.0374&-1.48&+0.1542&-0.0822&-1.16&+0.2359\\
        Ours w/o IB&-0.1698&-5.09&+0.1522&-0.1536&-0.60&+0.1412&-0.1031&-0.23&+0.0082&-0.0044&-1.19&+0.1178&-0.1712&-0.69&+0.1768\\
        Ours w/o CM and IB& -0.1765 & -5.93 & +0.1697 & -0.1578 & -0.71 & +0.1993 & -0.2087 & -1.06 & +0.2013 & -0.0951 & -2.07 & +0.2120 & -0.2005 & -1.52 & +0.2674 \\
        \bottomrule
    \end{tabular}
\end{table*}

To thoroughly evaluate the effectiveness of our proposed CSS pipeline in handling challenges such as occlusions, illumination variations, and sparse viewpoints in crowdsourced imagery, we conducted experiments across a diverse set of scenes. We selected several representative scenes from the Photo Tourism dataset \cite{image-matching-challenge-2024} (including Lincoln Memorial and Trevi Fountain) and from our own CSScenes dataset (Bingling Temple, Qianqing Palace and Yuanmingyuan Fountain). These scenes were specifically chosen for their inherent complexity, featuring significant occlusions, varying lighting conditions, and sparse, discontinuous viewpoints, making them well-suited for testing the robustness of our approach. The experiments were conducted on a server running Ubuntu 20.04.6 LTS with a 64-bit architecture. The system is powered by an Intel Xeon Platinum 8358P CPU @ 2.60GHz with 16 physical cores.For GPU acceleration, the system is outfitted with two NVIDIA A800-SXM4 GPUs.

Fig. \ref{fig:render-vis} illustrates a visual comparison of object appearance rendering across different methods, including ours, Wild Gaussian \cite{DBLP:journals/corr/abs-2407-08447}, Gaussian in the wild \cite{zhang2024gaussianwild3dgaussian}, CF-3DGS \cite{DBLP:journals/corr/abs-2312-07504}, and 3DGS\cite{10.1145/3569090}. Our method demonstrates superior texture and structural preservation under challenging lighting conditions and varied scene setups, particularly outperforming others in maintaining visual fidelity closer to the ground truth. Table \ref{tab:quant} quantitatively compares the rendering performance using SSIM \cite{1284395}, PSNR \cite{4775883}, and LPIPS \cite{DBLP:conf/cvpr/ZhangIESW18} metrics across multiple scenes. Our approach consistently achieves higher SSIM and lower LPIPS, indicating better structural integrity and visual quality. The ablation results further highlight the importance of crowd-sourced inputs and image variance components, where omitting these modules leads to noticeable performance degradation.

\section{Conclusion}

In this work, we introduced CSS, a pose-free 3D Gaussian Splatting framework that addresses key challenges of crowdsourced imagery, such as missing pose data and varying lighting. Through geometric priors and advanced illumination modeling, CSS consistently outperforms existing methods, as shown in experiments on the Photo Tourism and CSScenes datasets. While CSS offers an effective 3DGS-based reconstruction pipeline, especially for sparse and noisy data, it is only a step towards the broader goal of systematically restoring digital heritage. Achieving this requires a more robust, iterative crowdsourced system capable of continuous improvement. Future challenges include better handling of unstructured data and scaling to larger datasets, offering exciting opportunities for further advancement in 3D visual computing.

\bibliographystyle{IEEEbib}
\bibliography{strings,refs}

\begin{thebibliography}{10}

\bibitem{doi:10.1177/14780771231168224}
Nemeh Rihani,
\newblock ``Interactive immersive experience: Digital technologies for
  reconstruction and experiencing temple of bel using crowdsourced images and
  3d photogrammetric processes,''
\newblock {\em International Journal of Architectural Computing}, vol. 21, no.
  4, pp. 730--756, 2023.

\bibitem{22d9af9531054c88a8a46c194f6b3712}
Tino Mager and Carola Hein,
\newblock ``Digital excavation of mediatized urban heritage: Automated
  recognition of buildings in image sources,''
\newblock {\em Urban Planning}, vol. 5, no. 2, pp. 24--34, 2020.

\bibitem{Magnani_Douglass_Schroder_Reeves_Braun_2020}
Matthew Magnani, Matthew Douglass, Whittaker Schroder, Jonathan Reeves, and
  David~R. Braun,
\newblock ``The digital revolution to come: Photogrammetry in archaeological
  practice,''
\newblock {\em American Antiquity}, vol. 85, no. 4, pp. 737–760, 2020.

\bibitem{survey}
Ben Fei, Jingyi Xu, Rui Zhang, Qingyuan Zhou, Weidong Yang, and Ying He,
\newblock ``3d gaussian splatting as new era: A survey,''
\newblock {\em IEEE Transactions on Visualization and Computer Graphics}, pp.
  1--20, 2024.

\bibitem{3dgs}
Bernhard Kerbl, Georgios Kopanas, Thomas Leimk{\"{u}}hler, and George
  Drettakis,
\newblock ``3d gaussian splatting for real-time radiance field rendering,''
\newblock {\em {ACM} Trans. Graph.}, vol. 42, no. 4, pp. 139:1--139:14, 2023.

\bibitem{10630672}
Hao Wang and Minghui Li,
\newblock ``A new era of indoor scene reconstruction: A survey,''
\newblock {\em IEEE Access}, vol. 12, pp. 110160--110192, 2024.

\bibitem{10.1145/3569090}
Danzhao Cheng and Eugene Ch’ng,
\newblock ``Harnessing collective differences in crowdsourcing behaviour for
  mass photogrammetry of 3d cultural heritage,''
\newblock {\em J. Comput. Cult. Herit.}, vol. 16, no. 1, dec 2022.

\bibitem{csnerf}
Tong Qin, Changze Li, Haoyang Ye, Shaowei Wan, Minzhen Li, Hongwei Liu, and
  Ming Yang,
\newblock ``Crowd-sourced nerf: Collecting data from production vehicles for 3d
  street view reconstruction,''
\newblock {\em IEEE Transactions on Intelligent Transportation Systems}, pp.
  1--12, 2024.

\bibitem{colmapfree}
Yang Fu, Sifei Liu, Amey Kulkarni, Jan Kautz, Alexei~A. Efros, and Xiaolong
  Wang,
\newblock ``Colmap-free 3d gaussian splatting,''
\newblock in {\em Proceedings of the IEEE/CVF Conference on Computer Vision and
  Pattern Recognition (CVPR)}, June 2024, pp. 20796--20805.

\bibitem{10446844}
Wangze Xu, Qi~Wang, Xinghao Pan, and Ronggang Wang,
\newblock ``Hdpnerf: Hybrid depth priors for neural radiance fields from sparse
  input views,''
\newblock in {\em ICASSP 2024 - 2024 IEEE International Conference on
  Acoustics, Speech and Signal Processing (ICASSP)}, 2024, pp. 3695--3699.

\bibitem{wildinnerf}
Ricardo Martin-Brualla, Noha Radwan, Mehdi S.~M. Sajjadi, Jonathan~T. Barron,
  Alexey Dosovitskiy, and Daniel Duckworth,
\newblock ``Nerf in the wild: Neural radiance fields for unconstrained photo
  collections,''
\newblock in {\em 2021 IEEE/CVF Conference on Computer Vision and Pattern
  Recognition (CVPR)}, 2021, pp. 7206--7215.

\bibitem{9413356}
Jeong-Won HA, JUN-Sang YOO, and JONG-Ok KIM,
\newblock ``Deep color constancy using temporal gradient under ac light
  sources,''
\newblock in {\em ICASSP 2021 - 2021 IEEE International Conference on
  Acoustics, Speech and Signal Processing (ICASSP)}, 2021, pp. 2355--2359.

\bibitem{colmap1}
Johannes~L. Sch{\"{o}}nberger and Jan{-}Michael Frahm,
\newblock ``Structure-from-motion revisited,''
\newblock in {\em 2016 {IEEE} Conference on Computer Vision and Pattern
  Recognition, {CVPR} 2016, Las Vegas, NV, USA, June 27-30, 2016}. 2016, pp.
  4104--4113, {IEEE} Computer Society.

\bibitem{colmap2}
Johannes~L. Sch{\"{o}}nberger, Enliang Zheng, Jan{-}Michael Frahm, and Marc
  Pollefeys,
\newblock ``Pixelwise view selection for unstructured multi-view stereo,''
\newblock in {\em Computer Vision - {ECCV} 2016 - 14th European Conference,
  Amsterdam, The Netherlands, October 11-14, 2016, Proceedings, Part {III}},
  Bastian Leibe, Jiri Matas, Nicu Sebe, and Max Welling, Eds. 2016, vol. 9907
  of {\em Lecture Notes in Computer Science}, pp. 501--518, Springer.

\bibitem{DBLP:journals/corr/abs-2312-07504}
Yang Fu, Sifei Liu, Amey Kulkarni, Jan Kautz, Alexei~A. Efros, and Xiaolong
  Wang,
\newblock ``Colmap-free 3d gaussian splatting,''
\newblock {\em CoRR}, vol. abs/2312.07504, 2023.

\bibitem{fan2024instantsplat}
Zhiwen Fan, Wenyan Cong, Kairun Wen, Kevin Wang, Jian Zhang, Xinghao Ding,
  Danfei Xu, Boris Ivanovic, Marco Pavone, Georgios Pavlakos, Zhangyang Wang,
  and Yue Wang,
\newblock ``Instantsplat: Unbounded sparse-view pose-free gaussian splatting in
  40 seconds,'' 2024.

\bibitem{10446550}
Huachen Gao, Shihe Shen, Zhe Zhang, Kaiqiang Xiong, Rui Peng, Zhirui Gao,
  Qi~Wang, Yugui Xie, and Ronggang Wang,
\newblock ``Fdc-nerf: Learning pose-free neural radiance fields with flow-depth
  consistency,''
\newblock in {\em ICASSP 2024 - 2024 IEEE International Conference on
  Acoustics, Speech and Signal Processing (ICASSP)}, 2024, pp. 3615--3619.

\bibitem{Gao_2024_CVPR}
Xiangjun Gao, Xiaoyu Li, Chaopeng Zhang, Qi~Zhang, Yanpei Cao, Ying Shan, and
  Long Quan,
\newblock ``Contex-human: Free-view rendering of human from a single image with
  texture-consistent synthesis,''
\newblock in {\em Proceedings of the IEEE/CVF Conference on Computer Vision and
  Pattern Recognition (CVPR)}, June 2024, pp. 10084--10094.

\bibitem{DBLP:journals/jair/MogadalaKK21}
Aditya Mogadala, Marimuthu Kalimuthu, and Dietrich Klakow,
\newblock ``Trends in integration of vision and language research: {A} survey
  of tasks, datasets, and methods,''
\newblock {\em J. Artif. Intell. Res.}, vol. 71, pp. 1183--1317, 2021.

\bibitem{mast3r}
Vincent Leroy, Yohann Cabon, and J{\'{e}}r{\^{o}}me Revaud,
\newblock ``Grounding image matching in 3d with mast3r,''
\newblock {\em CoRR}, vol. abs/2406.09756, 2024.

\bibitem{Wang_2024_CVPR}
Shuzhe Wang, Vincent Leroy, Yohann Cabon, Boris Chidlovskii, and Jerome Revaud,
\newblock ``Dust3r: Geometric 3d vision made easy,''
\newblock in {\em Proceedings of the IEEE/CVF Conference on Computer Vision and
  Pattern Recognition (CVPR)}, June 2024, pp. 20697--20709.

\bibitem{10480646}
W.~Wang, R.~Luo, W.~Yang, and J.~Liu,
\newblock ``Unsupervised illumination adaptation for low-light vision,''
\newblock {\em IEEE Transactions on Pattern Analysis \& Machine Intelligence},
  vol. 46, no. 09, pp. 5951--5966, sep 2024.

\bibitem{otsu}
Nobuyuki Otsu,
\newblock ``A threshold selection method from gray-level histograms,''
\newblock {\em IEEE Transactions on Systems, Man, and Cybernetics}, vol. 9, no.
  1, pp. 62--66, 1979.

\bibitem{retinex}
Edwin~H Land,
\newblock ``The retinex theory of color vision,''
\newblock {\em Scientific american}, vol. 237, no. 6, pp. 108--129, 1977.

\bibitem{image-matching-challenge-2024}
Fabio Bellavia, Jiri Matas, Dmytro Mishkin, Luca Morelli, Fabio Remondino,
  Weiwei Sun, Amy Tabb, Eduard Trulls, Kwang~Moo Yi, Sohier Dane, and Ashley
  Chow,
\newblock ``Image matching challenge 2024 - hexathlon,'' 2024.

\bibitem{DBLP:journals/corr/abs-2407-08447}
Jonas Kulhanek, Songyou Peng, Zuzana Kukelova, Marc Pollefeys, and Torsten
  Sattler,
\newblock ``Wildgaussians: 3d gaussian splatting in the wild,''
\newblock {\em CoRR}, vol. abs/2407.08447, 2024.

\bibitem{zhang2024gaussianwild3dgaussian}
Dongbin Zhang, Chuming Wang, Weitao Wang, Peihao Li, Minghan Qin, and Haoqian
  Wang,
\newblock ``Gaussian in the wild: 3d gaussian splatting for unconstrained image
  collections,'' 2024.

\bibitem{1284395}
Zhou Wang, A.C. Bovik, H.R. Sheikh, and E.P. Simoncelli,
\newblock ``Image quality assessment: from error visibility to structural
  similarity,''
\newblock {\em IEEE Transactions on Image Processing}, vol. 13, no. 4, pp.
  600--612, 2004.

\bibitem{4775883}
Zhou Wang and Alan~C. Bovik,
\newblock ``Mean squared error: Love it or leave it? a new look at signal
  fidelity measures,''
\newblock {\em IEEE Signal Processing Magazine}, vol. 26, no. 1, pp. 98--117,
  2009.

\bibitem{DBLP:conf/cvpr/ZhangIESW18}
Richard Zhang, Phillip Isola, Alexei~A. Efros, Eli Shechtman, and Oliver Wang,
\newblock ``The unreasonable effectiveness of deep features as a perceptual
  metric,''
\newblock in {\em 2018 {IEEE} Conference on Computer Vision and Pattern
  Recognition, {CVPR} 2018, Salt Lake City, UT, USA, June 18-22, 2018}. 2018,
  pp. 586--595, Computer Vision Foundation / {IEEE} Computer Society.

\end{thebibliography}

\end{document}